# Tagging and Morphological Disambiguation of Turkish Text

Kemal Oflazer and İlker Kuruöz
Department of Computer Engineering and Information Science
Bilkent University
Bilkent, Ankara, TURKEY
{ko,kuruoz}@cs.bilkent.edu.tr

## Abstract

Automatic text tagging is an important component in higher level analysis of text corpora, and its output can be used in many natural language processing applications. In languages like Turkish or Finnish, with agglutinative morphology, morphological disambiguation is a very crucial process in tagging, as the structures of many lexical forms are morphologically ambiguous. This paper describes a POS tagger for Turkish text based on a full-scale two-level specification of Turkish morphology that is based on a lexicon of about 24,000 root words. This is augmented with a multi-word and idiomatic construct recognizer, and most importantly morphological disambiguator based on local neighborhood constraints, heuristics and limited amount of statistical information. The tagger also has functionality for statistics compilation and fine tuning of the morphological analyzer, such as logging erroneous morphological parses, commonly used roots, etc. Preliminary results indicate that the tagger can tag about 98-99% of the texts accurately with very minimal user intervention. Furthermore for sentences morphologically disambiguated with the tagger, an LFG parser developed for Turkish, generates, on the average, 50% less ambiguous parses and parses almost 2.5 times faster. The tagging functionality is not specific to Turkish, and can be applied to any language with a proper morphological analysis interface.

## 1 Introduction

As a part of large scale project on natural language processing for Turkish, we have undertaken the development of a number of tools for analyzing Turkish text. This paper describes one such tool – a text tagger for Turkish. The tagger is based on a full scale two-level morphological specification of Turkish (Oflazer, 1993), implemented on the PC-KIMMO environment (Antworth, 1990). In this paper, we describe the functionality and the performance of our tagger along with various techniques that we have employed to deal with various sources of ambiguities.

## 2 Tagging Text

Automatic text tagging is an important step in discovering the linguistic structure of large text corpora. Basic tagging involves annotating the words in a given text with various pieces of information, such as part-of-speech and other lexical features. Part-of-speech tagging facilitates higher-level analysis, such as parsing, essentially by performing a certain amount of ambiguity resolution using relatively cheaper methods.

The most important functionality of a tagger is the resolution of the structure and parts-of-speech of the lexical items in the text. This, however, is not a very trivial task since many words are in general ambiguous in their part-of-speech for various reasons. In English, for example a word such as *make* can be verb or a noun. In Turkish, even though there are ambiguities of such sort, the agglutinative nature of the language usually helps resolution of such ambiguities due to morphotactical restrictions. On the other hand, this very nature introduces another kind of ambiguity, where a lexical form can be morphologically interpreted in many ways. For example, the word *evin*, can be broken down as:[1]

|   | evin | POS | English |
|---|---|---|---|
| 1. | N(ev)+2SG-POSS | N | *(your) house* |
| 2. | N(ev)+GEN | N | *of the house* |
| 3. | N(evin) | N | *wheat germ* |

If, however, the local context is considered, it may be possible to resolve the ambiguity as in:

---

[1] Output of the morphological analyzer is edited for clarity.

| | sen-in | **ev-in** .. |
|---|---|---|
| | PN(you)+GEN | N(ev)+2SG-POSS |
| | *your* | *house* |

| | **evin** | kapı-sı .. |
|---|---|---|
| | N(ev)+GEN | N(door)+3SG-POSS |
| | *door* | *of the house* |

using genitive–possessive agreement constraints.

As a more complex case we can give the following:

**alınmış**
1. ADJ(al)+2SG-POSS+NtoV()+NARR+3SG[2]
   (V) *(it) was your red (one)*
2. ADJ(al)+GEN+NtoV()+NARR+3SG
   (V) *(it) belongs to the red (one)*
3. N(alın)+NtoV()+NARR+3SG
   (V) *(it) was a forehead*
4. V(al)+PASS+VtoAdj(mis)
   (ADJ) *(a) taken (object)*
5. V(al)+PASS+NARR+3SG
   (V) *(it) was taken*
6. V(alın)+VtoAdj(mis)
   (ADJ) *(an) offended (person)*
7. V(alın)+NARR+3SG
   (V) *(s/he) was offended*

It is in general rather hard to select one of these interpretations without doing substantial analysis of the local context, and even then one can not fully resolve such (usually semantic) ambiguities.

An additional problem that can be off-loaded to the tagger is the recognition of multi-word or idiomatic constructs. In Turkish, which abounds with such forms, such a recognizer can recognize these very productive multi-word constructs, like

| koş-a | koş-a |
|---|---|
| run+OPT+3SG | run+OPT+3SG |

| yap-ar | yap-ma-z |
|---|---|
| do+AOR+3SG | do+NEG+AOR+3SG |

where both components are verbal but the compound construct is a manner or temporal adverb. This relieves the parser from dealing with them at the syntactic level. Furthermore, it is also possible to recognize various proper nouns with this functionality. Such help from a tagging functionality would simplify the development of parsers for Turkish (Demir, 1993; Güngördü, 1993).

Researchers have used a number of different approaches for building text taggers. Karlsson (Karlsson, 1990) has used a rule-based approach where the central idea is to maximize the use of morphological information. Local constraints expressed as rules basically discard many alternative parses whenever possible. Brill (Brill, 1992) has designed a rule-based tagger for English. The tagger works by automatically recognizing rules and remedying its weaknesses, thereby incrementally improving its performance. More recently, there has been a rule-based approach implemented with finite-state machines (Koskenniemi et al., 1992; Voutilainen and Tapanainen, 1993).

A completely different approach to tagging uses statistical methods, (e.g., (Church, 1988; Cutting et al., 1993)). These systems essentially train a statistical model using a previously hand-tagged corpus and provide the capability of resolving ambiguity on the basis of most likely interpretation. The models that have been widely used assume that the part-of-speech of a word depends on the categories of the two preceding words. However, the applicability of such approaches to word-order free languages remains to be seen.

## 2.1 An example

We can describe the process of tagging by showing the analysis for the sentence:

*İşten döner dönmez evimizin yakınında bulunan derin gölde yüzerek gevşemek en büyük zevkimdi.*

*(Relaxing by swimming the deep lake near our house, as soon as I return from work was my greatest pleasure.)*

which we assume has been processed by the morphological analyzer with the following output:

| | işten | POS |
|---|---|---|
| 1. | N(iş)+ABL | N+ |
| | döner | |
| 1. | N(döner) | N |
| 2. | V(dön)+AOR+3SG | V+ |
| 3. | V(dön)+VtoAdj(er) | ADJ |
| | dönmez | |
| 1. | V(dön)+NEG+AOR+3SG | V+ |
| 2. | V(dön)+VtoAdj(mez) | ADJ |
| | evimizin | |
| 1. | N(ev)+1PL–POSS+GEN | N+ |
| | yakınında | |
| 1. | ADJ(yakın)+3SG–POSS+LOC | N+ |
| 2. | ADJ(yakın)+2SG–POSS+LOC | N |
| | bulunan | |
| 1. | V(bul)+PASS+VtoADJ(yan) | ADJ |
| 2. | V(bulun)+VtoADJ(yan) | ADJ+ |
| | derin | |
| 1. | N(deri)+2SG–POSS | N |
| 2. | ADJ(derin) | ADJ+ |
| 3. | V(der)+IMP+2PL | V |
| 4. | V(de)+VtoADJ(er)+2SG-POSS | N |
| 5. | V(de)+VtoADJ(er)+GEN | N |
| | gölde | |
| 1. | N(göl)+LOC | N+ |
| | yüzerek | |
| 1. | V(yüz)+VtoADV(yerek) | ADV+ |
| | gevşemek | |
| 1. | V(gevşe)+VtoINF(mak) | V+ |
| | en | |
| 1. | N(en) | N |
| 2. | ADV(en) | ADV+ |
| | büyük | |
| 1. | ADJ(büyük) | ADJ+ |
| | zevkimdi | |
| 1. | N(zevk)+1SG–POSS+ NtoV()+PAST+3SG | V+ |

---

[2]In Turkish, all adjectives can be used as nouns, hence with very minor differences adjectives have the same morphotactics as nouns.

Although there are a number of choices for tags for the lexical items in the sentence, almost all except one set of choices give rise to ungrammatical or implausible sentence structures.[3] There are number of points that are of interest here:

- the construct *döner dönmez* formed by two tensed verbs, is actually a temporal adverb meaning *... as soon as .. return(s)*, hence these two lexical items can be coalesced into a single lexical item and tagged as a temporal adverb.
- The second person singular possessive interpretation of *yakınında* is not possible since this word forms a simple compound noun phrase with the previous lexical item and the third person singular possessive morpheme functions as the compound marker, agreeing with the agreement of the previous genitive case-marked form.
- The word *derin (deep)* is the modifier of a simple compound noun *derin göl (deep lake)* hence the second choice can safely be selected. The verbal root in the third interpretation is very unlikely to be used in text, let alone in second person imperative form. The fourth and the fifth interpretations are not very plausible either. The first interpretation (meaning *your skin*) may be a possible choice but can be discarded in the middle of a longer compound noun phrase.
- The word *en* preceding an adjective indicates a superlative construction and hence the noun reading can be discarded.

## 3 The Tagging Tool

The tagging tool that we have developed integrates the following functionality with a user interface, as shown in Figure 1, implemented under X-windows. It can be used interactively, though user interaction is very rare and (optionally) occurs only when the disambiguation can not be done by the tagger.

1. Morphological analysis with error logging,
2. Multi-word and idiomatic construct recognition,
3. Morphological disambiguation by using constraints, heuristics and certain statistics,
4. Root and lexical form statistics compilation,

The second and the third functionalities are implemented by a rule-base subsystem which allows one to write rules of the following form:

$C_1:A_1; \quad C_2:A_2; \quad \ldots \quad C_n:A_n.$

where each $C_i$ is a set of constraints on a lexical form, and the corresponding $A_i$ is an action to be executed on the set of parses associated with that lexical form, *only when all the conditions are satisfied.*

---

[3]The correct choices of tags are marked with +.

The conditions refer to any available morphological or positional feature associated with a lexical form such as:

- Absolute or relative lexical position (e.g., sentence initial or final, or 1 after the current word, etc.)
- root and final POS category,
- derivation type,
- case, agreement (number and person), and certain semantic markers, for nominal forms,
- aspect and tense, subcategorization requirements, verbal voice, modality, and sense for verbal forms
- subcategorization requirements for postpositions.

Conditions may refer to absolute feature values or variables (as in Prolog, denoted by the prefix _ in the following examples) which are then used to link conditions. *All occurrences of a variable have to unify for the match to be considered successful.* This feature is powerful and and lets us specify in a rather general way, (possibly long distance) feature constraints in complex NPs, PPs and VPs. This is a part of our approach that distinguishes it from other constraint-based approaches.

The actions are of the following types:

- **Null action**: Nothing is done on the matching parse.
- **Delete**: Removes the matching parse if more than one parse for the lexical form are still in the set associated with the lexical form.
- **Output**: Removes all but the matching parse from the set effectively tagging the lexical form with the matching parse.
- **Compose**: Composes a new parse from various matching parses, for multi-word constructs.

These rules are ordered, and applied in the given order and actions licensed by any matching rule are applied. One rule formalism is used to encode both multi-word constructs and constraints.

### 3.1 The Multi-word Construct Processor

As mentioned before, tagging text on lexical item basis may generate spurious or incorrect results when multiple lexical items act as single syntactic or semantic entity. For example, in the sentence *Şirin mi şirin bir köpek koşa koşa geldi* (A very cute dog came running) the fragment *şirin mi şirin* constitutes a duplicated emphatic adjective in which there is an embedded question suffix *mi* (written separately in Turkish),[4] and the fragment *koşa koşa* is a duplicated verbal construction, which has the grammatical role of manner adverb in the sentence, though both of the constituent forms are verbal constructions. The purpose of the multi-word construct processor is to detect and tag such productive constructs in addition to various other semantically coalesced forms such as proper nouns, etc.

---

[4]If, however, the adjective *şirin* was not repeated, then we would have a question formation.

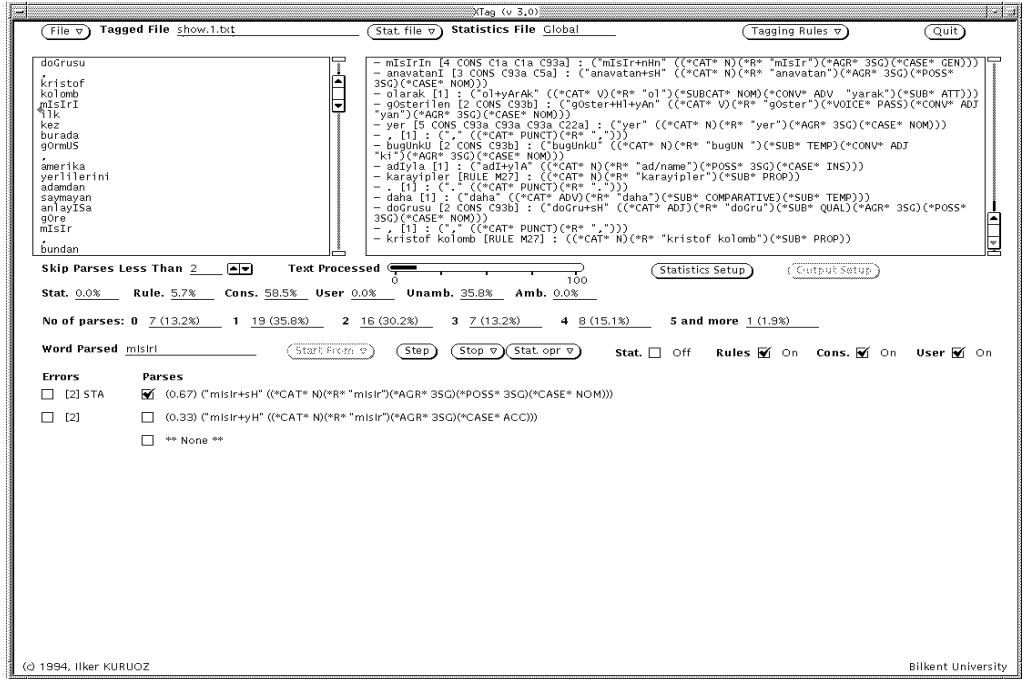

Figure 1: User interface of tagging tool

The following is a set of multi-word constructs for Turkish that we handle in our tagger. This list is not meant to be comprehensive, and new construct specifications can easily be added. It is conceivable that such a functionality can be used in almost any language.

1. duplicated optative and 3SG verbal forms functioning as manner adverb, e.g., *koşa koşa*, aorist verbal forms with root duplications and sense negation functioning as temporal adverbs, e.g., *yapar yapmaz*, and duplicated verbal and derived adverbial forms with the same verbal root acting as temporal adverbs, e.g., *gitti gideli*,

2. duplicated compound nominal form constructions that act as adjectives, e.g., *güzeller güzeli*, and emphatic adjectival forms involving the question suffix, e.g., *güzel mi güzel*,

3. adjective or noun duplications that act as manner adverbs, e.g., *hızlı hızlı*, *ev ev*,

4. idiomatic word sequences with specific usage whose semantics is not compositional, e.g., *yanı sıra*, *hiç olmazsa*, and idiomatic forms which are never used singularly, e.g., *gürül gürül*,

5. proper nouns, e.g., *Jimmy Carter*, *Topkapı Sarayı* (Topkapı Palace).

6. compound verb formations which are formed by a lexically adjacent, direct or oblique object and a verb, which for the purposes of syntactic analysis, may be considered as single lexical item.

We can give the following example for specifying a multi-word construct:[5]

```
Lex=_W1, Root=_R1, Cat=V, Aspect=AOR, Agr=3SG,
       Sense=POS: ;
Lex=_W2, Root=_R1, Cat=V, Aspect=AOR, Agr=3SG,
       Sense = NEG:
Compose=((*CAT* ADV)(*R* "_W1 _W2 (_R1)")
                             (*SUB* TEMP)).
```

This rule would match any adjacent verbal lexical forms with the same root, both with the aorist aspect, and 3SG agreement. The first verb has to be positive and the second one negated. When found, a composite lexical form with an temporal adverb part-of-speech, is then generated. The original verbal root may be recovered from the root of the composed form for any subcategorization checks, at the syntactic level.

### 3.2 Using constraints for morphological ambiguity resolution

Morphological analysis does not have access to syntactic context, so when the morphological structure of a lexical form has several distinct analyses, it is not possible to disambiguate such cases except maybe by using root usage frequencies. For disambiguation one may have to use information provided by sentential position and the local morphosyntactic context. Voutilainen and Heikkila (Voutilainen et al., 1992) have proposed a *constraint grammar* approach where one specifies constraints on the local context of a word to disambiguate among multiple readings of a word. Their approach has, however, been applied to English where morphological information has rather little use in such resolution.

---

[5]The output of the morphological analyzer is actually a *feature-value* list in the standard LISP format.

In our tagger, constraints are applied on each word, and check if the forms within a specified neighborhood of the word satisfy certain morphosyntactic or positional restrictions, and/or agreements. Our constraint pattern specification is very similar to multi-word construct specification. Use of variables, operators and actions, are same except that the compose actions does not make sense here. The following is an example constraint that is used to select the postpositional reading of certain word when it is preceded by a yet unresolved nominal form with a certain case. The only requirement is that the case of the nominal form agrees with the case subcategorization requirement of the following postposition. (LP = 0 refers to current word, LP = 1 refers to next word.)

```
LP = 0, Case = _C : Output;
LP = 1, Cat = POSTP, Subcat = _C : Output.
```

When a match is found, the matching parses from both words are selected and the others are discarded. This one constraint disambiguates almost all of the postpositions and their arguments, the exceptions being nominal words which semantically convey the information provided by the case (such as words indicating direction, which may be used as if they have a dative case).

Finally the following example constraint deletes the sentence final adjectival readings derived from verbs, effectively *preferring* the verbal reading (as Turkish is a SOV language.)

```
Cat = V, Finalcat = ADJ, SP = END : Delete.
```

## 4 Performance of the Tagger

We have performed some preliminary experiments to assess the effectiveness of our tagger. We have used about 250 constraints for Turkish. Some of these constraints are very general as the postposition rule above, while some are geared towards recognition of NP's of various sorts and a small number apply certain syntactic heuristics. In this section, we summarize our preliminary results. Table 1 presents some preliminary results about the our tagging experiments.

Although the texts that we have experimented with are rather small, the results indicate that our approach is effective in disambiguating morphological structures, and hence POS, with minimal user intervention. Currently, the speed of the tagger is limited by essentially that of the morphological analyzer, but we have ported the morphological analyzer to the XEROX TWOL system developed by Karttunen and Beesley (Karttunen and Beesley, 1992). This system can analyze Turkish word forms at about 1000 forms/sec on SparcStation 10's. We intend to integrate this to our tagger soon, improving its speed performance considerably.

We have tested the impact of morphological disambiguation on the performance of a LFG parser developed for Turkish (Güngördü, 1993; Güngördü and Oflazer, 1994). The input to the parser was disambiguated using the tool developed and the results were compared to the case when the parser had to consider all possible morphological ambiguities itself. For a set of 80 sentences considered, it can be seen that (Table 2), morphological disambiguation enables almost a factor of two reduction in the average number of parses generated and over a factor of two speed-up in time.

## 5 Conclusions

This paper has presented an overview of a tool for tagging text along with various issues that have come up in disambiguating morphological parses of Turkish words. We have noted that the use of constraints is very effective in morphological disambiguation. Preliminary results indicate that the tagger can tag about 98-99% of the texts accurately with very minimal user intervention, though it is conceivable that it may do worse on more substantial text – but there is certainly room for improvement in the mechanisms provided. The tool also provides for recognition of multi-word constructs that behave as a single syntactic and semantic entity in higher level analysis, and the compilation of information for fine-tuning of the morphological analyzer and the tagger itself. We, however, feel that our approach does not deal satisfactorily with most aspects of word-order freeness. We are currently working on an extension whereby the rules do not apply immediately but vote on their preferences and a final global vote tally determines the assignments.

## 6 Acknowledgment

This research was supported in part by a NATO Science for Stability Program Grant, TU-LANGUAGE.

Table 1: Statistics on texts tagged, and tagging and disambiguation results

| Text | Words | Morphological Parse Distribution | | | | | |
|---|---|---|---|---|---|---|---|
| | | 0 | 1 | 2 | 3 | 4 | ≥ 5 |
| 1 | 468 | 7.3% | 28.7% | 41.1% | 11.1% | 7.1 % | 4.7% |
| 2 | 573 | 1.0% | 30.2% | 37.3% | 13.1% | 11.1% | 7.3% |
| 3 | 533 | 3.8% | 24.8% | 38.1% | 19.1% | 9.2 % | 5.0% |
| 4 | 7004 | 3.9% | 17.2% | 41.5% | 15.6% | 11.7% | 10.1% |

Note: Words with zero parses are proper names which are not in the lexicon of the morphological analyzer.

| Text | % Correctly Tagged Automatically | % Tagged by User | % Correctly Tagged Total | Automatic Disambiguation by | |
|---|---|---|---|---|---|
| | | | | Multi-word Rules | Constraints |
| 1 | 98.5 | 1.0 | 99.1 | 10.1 | 67.7 |
| 2 | 98.5 | 0.3 | 98.8 | 7.5 | 74.4 |
| 3 | 97.8 | 1.1 | 98.9 | 3.1 | 74.5 |
| 4 | 95.4 | 1.7 | 97.1 | 4.2 | 76.4 |

Table 2: Impact of disambiguation on parsing performance

| | No disambiguation | | With disambiguation | | Ratios | |
|---|---|---|---|---|---|---|
| Avg. Length (words) | Avg. parses | Avg. time (sec) | Avg. parses | Avg. time (sec) | parses | speed-up |
| 5.7 | 5.78 | 29.11 | 3.30 | 11.91 | 1.97 | 2.38 |

Note: The ratios are the averages of the sentence by sentence ratios.